\documentclass[10pt,twocolumn,letterpaper]{article}

\usepackage{cvpr}              % To produce the CAMERA-READY version
\usepackage{pifont}
\definecolor{cvprblue}{rgb}{0.21,0.49,0.74}
\usepackage[pagebackref,breaklinks,colorlinks,allcolors=cvprblue]{hyperref}

\usepackage{graphicx}
\usepackage{amsmath}
\usepackage{amssymb}
\usepackage{booktabs}
\usepackage{amsmath} 
\usepackage{graphicx}
\newcommand{\name}{PanoEnv}

\def\paperID{13556} % *** Enter the Paper ID here
\def\confName{CVPR}
\def\confYear{2026}

\title{PanoEnv: Exploring 3D Spatial Intelligence in Panoramic Environments with Reinforcement Learning}

\author{
Zekai Lin\\
University of Glasgow\\
\and
Xu Zheng$^{*}$\\
HKUST(GZ)\\
{\small $^{*}$Corresponding author.}
\and
\small{Code \& Dataset: \url{https://github.com/7zk1014/PanoEnv}.}
}

\usepackage{tikz}
\usetikzlibrary{positioning,arrows.meta,calc}

\usepackage{booktabs}   % 更漂亮的表格线
\usepackage{multirow}   % 多行合并
\usepackage{colortbl}   % 灰底
\usepackage{xcolor}     % 定义颜色
\usepackage{graphicx}   % 可调整表格大小
\usepackage{xcolor}
\usepackage{amsmath}

\usepackage{float}      % 调整图片位置

\usepackage{caption}
\usepackage{subcaption} % ← 加入导言区

\begin{document}

\twocolumn[{
\maketitle
% \vspace{-2.9em} % 标题与图之间压一点
\begin{center}
% 等比例缩小，使用 width 控制
\includegraphics[width=\textwidth]{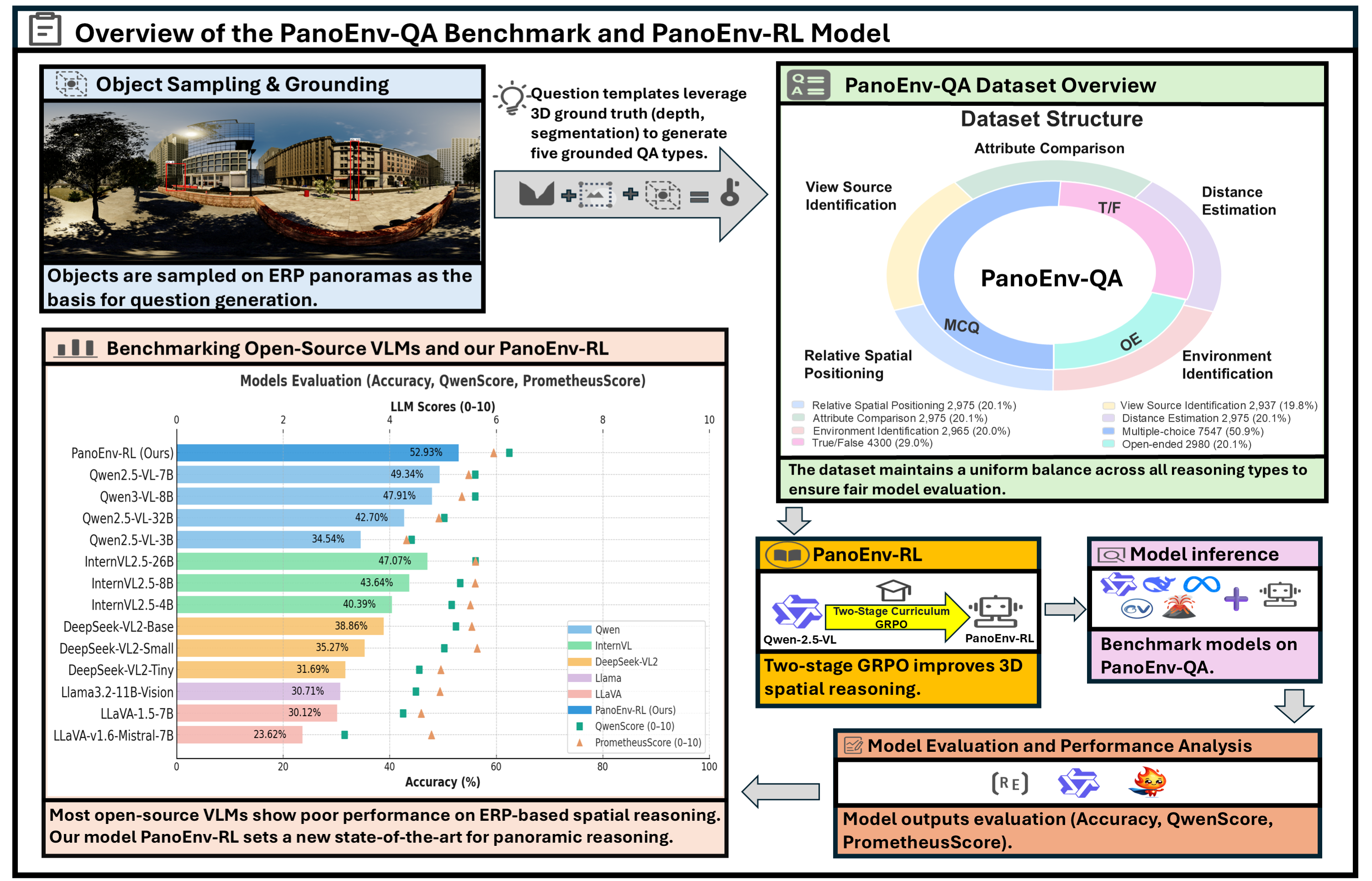} % ,height=0.47\textheight
\end{center}
% \vspace{-2.0em} % 图与 caption 之间
\begin{center}
\small
\textbf{Figure 1.} Overview of the PanoEnv framework, including the PanoEnv-QA benchmark and the RL-enhanced PanoEnv-RL model.
\label{fig:teaser}
\end{center}
\vspace{0.6em} % caption 和 abstract 之间
}]

\begin{abstract}
360° panoramic images are increasingly used in VR, autonomous driving, and robotics for holistic scene understanding. However, current Vision–Language Models (VLMs) struggle with 3D spatial reasoning on Equirectangular Projection (ERP) images due to geometric distortion and limited 3D supervision. We introduce \textbf{\textit{PanoEnv}}, a large-scale VQA benchmark built from synthetic 3D environments, containing 14.8K questions across five categories (e.g., relative position, volume comparison) grounded in accurate 3D annotations—depth, segmentation, and bounding boxes. Benchmarking 14 state-of-the-art VLMs reveals limited 3D understanding, achieving only 49.34\% overall and 8.36\% on open-ended (OE) questions. To enhance 3D reasoning, we propose a reinforcement learning post-training framework based on Group Relative Policy Optimization (GRPO) with a ground-truth-guided reward combining five geometry-aware strategies (e.g., distance tolerance, spatial consistency). A two-stage curriculum further mitigates catastrophic forgetting: Stage\~1 trains on structured tasks (T/F, MCQ), and Stage\~2 fine-tunes on mixed OE data for generalization. Our 7B model sets a new SoTA performance, improving total accuracy to 52.93\% (+3.59\%) and OE accuracy to 14.83\% while maintaining structured-task performance. It also achieves top semantic scores (Q-Score 6.24, P-Score 5.95), surpassing 32B models. These results demonstrate that PanoEnv-QA and our curriculum-based RL framework effectively instill 3D spatial intelligence in VLMs for omnidirectional perception.
\end{abstract}
    
\section{Introduction}
\label{sec:intro}
Omnidirectional images (ODIs) capture a full $360^\circ$ field of view in a single panorama, providing holistic scene coverage essential for VR/AR, autonomous systems, and embodied agents~\cite{ai2025survey,zheng2025panorama,zheng2023both,zheng2023look,zheng2024semantics,zheng2024360sfuda++,zhang2024goodsam,zhu2024customize,zhong2025omnisam,liao2025memorysam,zhang2024goodsam++}. Unlike pinhole images, ODIs retain long-range context and off-axis relations, enabling global awareness, navigation, and safety assessment. Recent Vision--Language Models (VLMs) show strong generalization across diverse visual--text modalities~\cite{radford2021learning,dai2023instructblip,chen2024internvl,li2024llava,wang2024qwen2,li2022blip,caffagni2024revolution}, yet when ODIs are represented via Equirectangular Projection (ERP), they face two key challenges: complex scene geometry and limited 3D supervision. These issues distort perception (e.g., scale and shape vary with latitude) and hinder reasoning (e.g., 3D relations map nonlinearly to 2D ERP coordinates)~\cite{cai2024interact360,zheng2025panorama,cao2025unlocking,zhang2025omnidirectional}.

Spatial questions that appear simple on perspective images—“Is object A left of B?”, “How far is A?”, “Which is larger in 3D?”—become nontrivial on ERP panoramas. First, ERP stretches pixels near the poles, distorting cues learned from pinhole images. Second, ODIs require reasoning over long-range dependencies, as each panorama spans multiple disjoint views with discontinuities at seams. Third, existing VQA datasets~\cite{antol2015vqa,goyal2017vqa2,wu2017visual,yun2021pano} rarely include physical ground truth (e.g., depth, 3D boxes, precise geometry) aligned with ODIs, limiting the evaluation and learning of 3D reasoning. Prior panoramic VQA and omnidirectional reasoning benchmarks~\cite{chou2020visual,dongfang2025multimodal,zhang2025towards,yang2025odi,li2025viewspatial} are valuable early efforts but still lack the combination of panoramic coverage, dense geometry, and fine-grained, geometry-grounded supervision needed for reinforcement learning.

To this end, we introduce \textbf{\name}, a large-scale VQA benchmark designed specifically to probe 3D spatial intelligence on ERP panoramas. Built from synthetic but photo-realistic 3D environments, \name-QA offers over 14.8K questions spanning five categories that progressively require stronger 3D understanding: camera view source identification, environment identification, distance estimation, relative spatial positioning in 3D, and intrinsic attribute comparison (true size/shape)---all grounded in precise 3D annotations (depth, semantics, and 3D bounding boxes). Because questions and answers are programmatically derived from physical ground truth, \name-QA can serve both as a reliable evaluation bed and as a source of \emph{verifiable} supervision signals.

We evaluate 14 state-of-the-art VLMs~\cite{chen2024internvl,li2024llava,liu2024improved,bai2025qwen2,yang2025qwen3,chen2024expanding,wu2024deepseek,dubey2024llama} on a challenging subset of \name-QA and find that overall performance is low (best 49.34\% accuracy), with a near collapse on open-ended (OE) questions (best 8.36\%). While some models handle binary or multiple-choice prompts moderately well, they struggle to generate precise, geometry-consistent responses that require metric reasoning (e.g., distances) or composing multi-axis relations (e.g., ``above and to the left of''). This behavior suggests that current training paradigms bias toward 2D heuristics rather than building a faithful 3D scene representation from ERP imagery.

To bridge this gap, we propose a 3D-aware RL post-training framework based on Group Relative Policy Optimization (GRPO)~\cite{shao2024deepseekmath}. Two key designs underpin our approach. First, we introduce a ground-truth-guided routed reward with five specialized strategies aligned to question types (e.g., numerical tolerance for distance, axis-wise matching for spatial relations). This replaces noisy proxy rewards (e.g., LLM judges) with supervision anchored in physical reality. Second, we employ a two-stage curriculum: Stage~1 trains on structured tasks (T/F, MCQ) to stabilize policy learning; Stage~2 reintroduces OE data to recover generative skills while mitigating catastrophic forgetting. Together, these designs significantly improve overall accuracy and OE reasoning quality.

Starting from a 7B VLM backbone, our RL-trained model achieves a new SoTA performance on \name-QA (52.93\% accuracy), surpassing much larger 26B/32B baselines. Most notably, OE accuracy improves from 6.39\% to \textbf{14.83\%} (+132\% relative), with minimal degradation on structured tasks. These gains demonstrate that (1) ERP-specific geometric challenges can be addressed when training signals are physically grounded, and (2) curriculum design is crucial for balancing discrete decision-making and free-form generation.

Overall, our work makes three contributions:
\ding{172} \textbf{\name:} a large-scale, geometry-grounded panoramic VQA benchmark for 3D spatial intelligence on ERP images. It offers five complementary categories with pixel/point-aligned physical annotations, enabling both faithful evaluation and RL supervision.
\ding{173} \textbf{Comprehensive benchmarks of 14 VLMs:} we show consistent failures on 3D reasoning, especially for open-ended generation, revealing a significant gap between current training paradigms and panoramic spatial understanding.
\ding{174} \textbf{3D-aware Curriculum-based RL post-training:} a GRPO-based framework with routed, ground-truth rewards and a two-stage curriculum that sets a new SOTA with a 7B model and yields large relative gains on OE questions. A high-level overview of the dataset design, model evaluation, and our RL framework is provided in the teaser figure (Fig.~1).
\section{Related Work}
\label{sec:relatedwork}

\noindent \textbf{VQA and Reasoning Datasets.}
Visual Question Answering (VQA) is a core benchmark for machine intelligence~\cite{zhang2025egonight,zhu2026egosound,li2025egocross,pan2026v2sam,fu2025objectrelator,zheng2025multimodal,li2025chorus,shu2025earthmind}. Early datasets mainly tested recognition abilities, such as identifying object attributes or counts. The original VQA dataset~\cite{antol2015vqa} established this paradigm, but later studies showed that models often exploit language priors instead of genuine image understanding. VQA v2.0~\cite{goyal2017vqa2} addressed this issue by balancing question distributions. Subsequent work shifted toward datasets requiring higher-level reasoning rather than simple pattern matching. Datasets like GQA~\cite{hudson2019gqa} and CLEVR~\cite{johnson2017clevr} emphasize spatial and logical relationships between objects, while VCR~\cite{zellers2019vcr} further stresses commonsense reasoning by asking models to justify their answers. More recent benchmarks such as What’s-Up~\cite{kamath2023s} and VSI-Bench~\cite{yang2025thinking} explore spatial understanding and cognitive mapping. Despite advances, most reasoning datasets remain confined to 2D perspective images and compositional or commonsense logic. Although several datasets~\cite{chou2020visual,dongfang2025multimodal,fan2019egovqa} have emerged, they are limited to single-agent or egocentric scenarios rather than full-scene spatial reasoning. These gaps motivate our work: a new benchmark for inferring 3D physical relationships from 2D panoramas.

\noindent \textbf{Omnidirectional VQA and Spatial Reasoning.}
The complete field of view in 360° images makes them valuable for immersive applications and embodied intelligence ~\cite{ai2022deep}. However, existing MLLMs struggle to interpret the geometric distortions and complex spatial relationships inherent in these images. Early research such as VQA 360°~\cite{chou2020visual} extended VQA to panoramic images, but the dataset is not publicly released. Two recent works, OSR-Bench~\cite{dongfang2025multimodal} and 360-R1~\cite{zhang2025towards}, have significantly advanced the field. OSR-Bench established the first large-scale panoramic spatial reasoning benchmark, systematically documenting the drawbacks of state-of-the-art models. 360-R1 created the OmniVQA dataset and applied reinforcement learning to enhance panoramic reasoning capabilities. Our research builds upon these pioneering works to address remaining gaps. It introduces \textbf{\textit{higher-order, multi-object relational questions}}, including:
\textbf{\textit{intrinsic attribute comparison}} (e.g., real 3D volume and shape),
\textbf{\textit{3D position judgment}},
\textbf{\textit{depth comparison}},
\textbf{\textit{surrounding environment analysis}}, and
a unique meta-reasoning challenge: \textbf{\textit{original source viewpoint judgment}}.
Furthermore, a ``2D textual reference, 3D spatial query'' paradigm is employed. This method uses an object's 2D attributes to generate a reference, then queries the model about its relationships in 3D space, thereby constituting a difficult cross-dimensional reasoning task.

% \noindent \textbf{RL for Multimodal Reasoning.}
% Supervised fine-tuning (SFT) is the standard training method for MLLMs, but its reasoning processes can be rigid~\cite{dong2024abilities}. RL offers a promising alternative, as demonstrated by pioneering works~\cite{guo2025deepseek,ouyang2022training}, such as Vision-R1~\cite{huang2025vision}, which converts visual inputs into structured reasoning chains. In the panoramic VQA domain, 360-R1~\cite{zhang2025towards} also applies an RL framework with a reward function based on semantic similarity. Previous methods rely on other LLMs or human annotations to get answers or CoT, which introduce biases or errors. The core innovation in this research is the use of geometric ground truth from TartanAir~\cite{wang2020tartanair} to \textbf{\textit{programmatically generate correct textual answers}}, which then serve as the basis for reward calculation. During RL training, the reward is based on the correspondence between the model's output and this physically-grounded ground truth text. This method ensures the learning signal originates entirely from the objective physical world, providing a more direct incentive for the model to learn 3D spatial structures.

% =========================
% 3. Methodology
% =========================
\section{Methodology}
\label{sec:method}

In this section, we present the construction of the PanoEnv-QA dataset and our 3D-aware RL post-training framework, including the GRPO optimization scheme, a multi-faceted reward system, and a two-stage curriculum.

% ---------------- 3.1 Dataset ----------------
\subsection{PanoEnv-QA Construction}
\label{sec:dataset_construction}

\setcounter{figure}{1}
\begin{figure*}[t]
\centering
\includegraphics[width=0.99\textwidth]{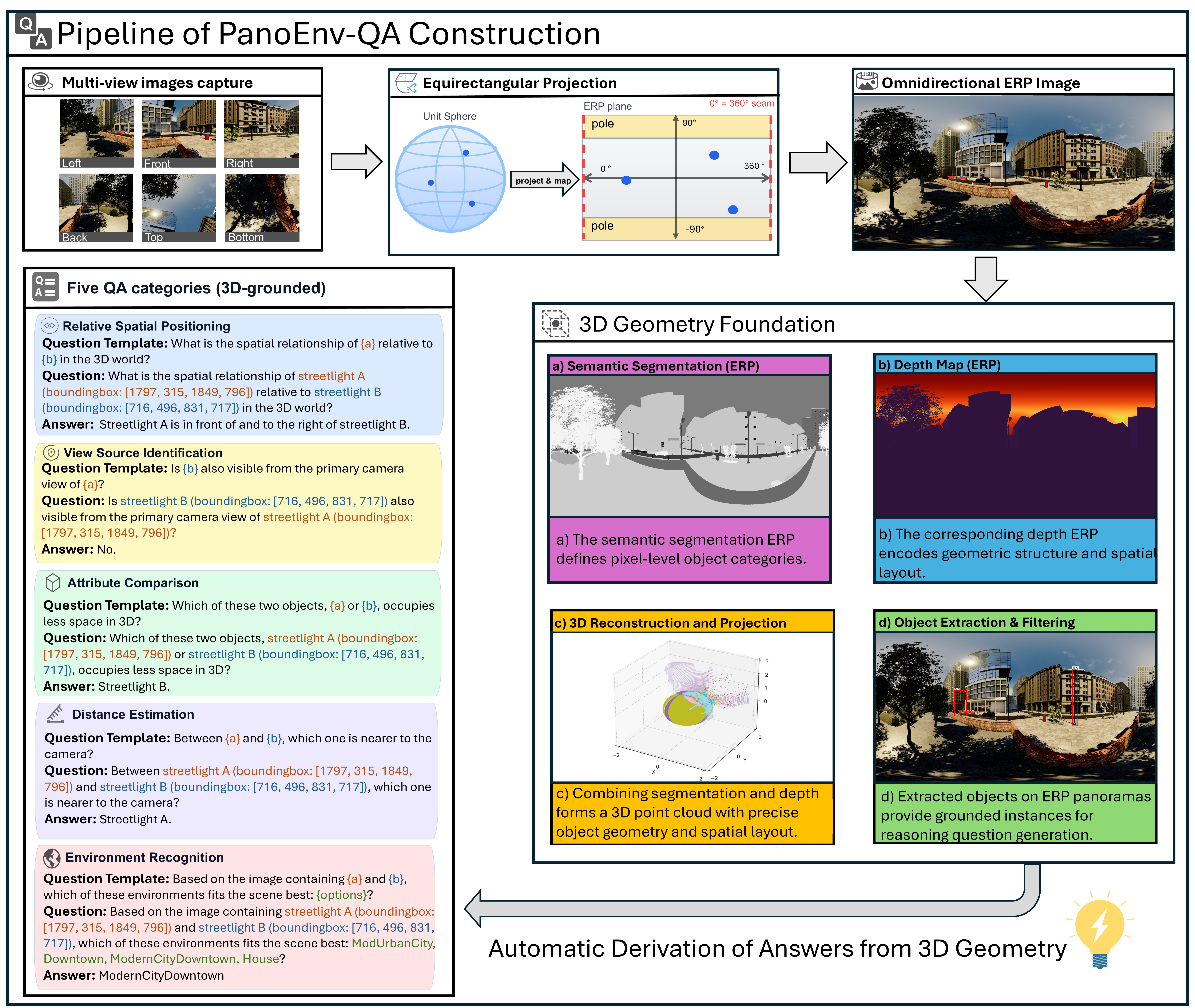}
% \vspace{-4pt}
\caption{
\textbf{Overview of the PanoEnv-QA construction pipeline.}
We convert multi-view TartanAir data into ERP panoramas and generate geometry-grounded QA pairs using depth, semantics, and 3D projections.
}
% \vspace{-6pt}
\label{fig:pipeline}
\end{figure*}

\subsubsection{Overview and Data Foundation}
\label{sec:dataset_overview}

Our goal is to construct a large-scale, multi-modal QA dataset to enhance and evaluate the spatial intelligence of MLLMs in 360° panoramic environments across indoor, outdoor, and synthetic scenes. The benchmark is designed to push models beyond 2D pattern recognition toward genuine 3D reasoning about scene geometry and object relations. We build upon TartanAir~\cite{wang2020tartanair}, a synthetic dataset chosen for its precise 3D ground truth (depth and segmentation), which is difficult to obtain in real-world data. For each scene, six perspective views from the cubemap camera setup are merged into a high-resolution ERP image. Each ERP instance provides perfectly aligned RGB, depth, and semantic segmentation modalities, ensuring pixel-level correspondence between appearance, geometry, and semantics. 
As illustrated in Fig.~\ref{fig:pipeline}, our dataset construction pipeline transforms multi-view TartanAir data into ERP panoramas and structured QA pairs grounded in 3D geometry.

For systematically evaluation, our QA generation focuses on five distinct yet complementary categories. These range from foundational awareness of the ERP format (\textit{Camera View Source Identification}) and geometric quantification (\textit{Object Distance Estimation}) to high-level context (\textit{Environment Identification}). The most challenging categories require true 3D reasoning about \textit{Relative Spatial Positioning} and an understanding of objects' intrinsic physical properties through \textit{Intrinsic Attribute Comparison}.

\subsubsection{Question-Answer Pair Generation}
\label{sec:qa_generation}

PanoEnv generation pipeline automatically leverages ground-truth data to produce diverse and challenging QA pairs. For each panoramic scene, we conduct an object-centric analysis using semantic segmentation maps to identify and instance all visible objects. To ensure relevance, we filter out small objects and amorphous background categories (e.g., \texttt{sky}, \texttt{ground}, \texttt{wall}). For comparative questions, object pairs with significant 2D bounding box containment are excluded to avoid trivial or ambiguous cases (e.g., comparing a tire to the car it belongs to). For each valid instance $O_i$, we extract high-fidelity properties directly from its segmentation mask $M_i$, including 2D bounding box $B_i$, depth statistics, camera source, 3D point cloud, and computed volume. 
Followings are the five templates: 

\noindent \textbf{\ding{172} Camera View Source Identification.}
This category evaluates whether the model recognizes that an ERP image is a composite panorama stitched from multiple perspective views, rather than a single photo. Understanding this structure is essential for handling artifacts near seam boundaries and reasoning about scene continuity beyond limited fields of view. The geometric mapping from the 2D ERP to the six cubemap views provides precise ground truth. For any pixel $(p_x, p_y)$ in an ERP image of size $W \times H$, the corresponding spherical coordinates are:
\begin{equation}
\setlength{\abovedisplayskip}{3pt}
\setlength{\belowdisplayskip}{3pt}
\label{eq:pixel_to_spherical}
\lambda = \left(\frac{p_x}{W} - 0.5\right) 2\pi, \quad
\phi = -\left(\frac{p_y}{H} - 0.5\right) \pi
\end{equation}
These define a 3D unit direction vector whose dominant axis determines the source view (e.g., \textit{front}, \textit{left}, \textit{top}).
For each object, we sample $N=100$ points from its segmentation mask $M_i$ to determine its \textit{visible cameras}. Objects seen from multiple views are labeled as \textit{seam objects}. QA pairs are then generated from these attributes.

\noindent \textbf{\ding{173} Object Distance Estimation.}
This category evaluates a model’s ability to perform quantitative and qualitative depth reasoning, moving beyond 2D heuristics (e.g., size as a proxy for distance) toward true 3D understanding. Ground truth depths are derived from the ERP depth map $D_{ERP}$. For each object $O_i$, we extract valid depth values $\mathcal{D}_i$ within its segmentation mask $M_i$, ensuring exclusion of background and occlusions. From this distribution, we compute key statistics—median ($p_{50}$), percentiles, and inter-quartile range (IQR)—to obtain a robust depth profile. (More details refer to Suppl.Mat.)

\noindent \textbf{\ding{174} Environment Identification.}
This category evaluates high-level scene understanding and contextual reasoning, testing whether the model can classify environments based on object composition and architectural style in the 360° panorama. We adopt a metadata-driven approach using TartanAir’s ground-truth environment labels, organized into a two-level taxonomy: \textit{Scene Attribute} (e.g., \textit{indoor}, \textit{outdoor}) and \textit{Scene Category} (e.g., \textit{Urban}, \textit{Nature}). For multiple-choice questions, distractors are selected semantically rather than randomly—e.g., pairing a “day” scene with its “night” or “winter” variant—to encourage fine-grained reasoning beyond simple keyword cues.

\noindent \textbf{\ding{175} Relative Spatial Positioning.}
This category assesses a model’s ability to reconstruct accurate 3D spatial relationships between objects—an inherently difficult task due to ERP distortions. We achieve this by inversely projecting object locations from the ERP image to a unified 3D Cartesian space. The coordinate system is right-handed, with +Y upward, +X rightward, and –Z forward (aligned with the front-facing view at $\lambda=0$).
For determining an object's single 3D locator, we use the centroid $(p_x, p_y)$ of its 2D bounding box and its median depth $d_i$. We first compute its spherical coordinates $(\lambda_i, \phi_i)$ using Eq.~\ref{eq:pixel_to_spherical}. We then convert this spherical representation into a 3D centroid coordinate $\vec{P}_i = [x_i, y_i, z_i]^T$ using the standard transformation:
\begin{equation}
\setlength{\abovedisplayskip}{3pt}
\setlength{\belowdisplayskip}{3pt}
\label{eq:spherical_to_cartesian}
\begin{aligned}
    x_i &= -d_i \cdot \cos(\phi_i) \cdot \sin(\lambda_i), % <-- [CRITICAL FIX] Added minus sign
     y_i = d_i \cdot \sin(\phi_i), \\
    z_i &= -d_i \cdot \cos(\phi_i) \cdot \cos(\lambda_i)
\end{aligned}
\end{equation}
The spatial relation between two objects $O_i$ and $O_j$ is defined as $\vec{V}_{ij} = \vec{P}i - \vec{P}j$. Linguistic labels (e.g., “above,” “in front of”) are obtained by comparing $\vec{V}{ij}$ components with a threshold $\tau{\text{pos}}$. This geometry-based approach ensures all QA pairs are grounded in precise 3D relationships.

\noindent \textbf{\ding{176} Intrinsic Attribute Comparison.}
To probe the model's understanding of intrinsic, view-independent physical properties of objects. This category moves beyond spatial location to assess reasoning about 3D shape and size, requiring the model to infer physical characteristics from 2D projections and depth information. The ground truth for this category is derived from the 3D representation of each object, which is constructed by projecting every pixel within its segmentation mask $M_i$ into 3D space. From this accurate point cloud, we compute a tight 3D bounding box and its corresponding physical dimensions (length, width, height). These properties allow us to generate two types of comparative questions: \ding{172} \textbf{Volume Comparison:} We calculate the precise volume of the 3D bounding box for each object. This allows us to generate questions asking which of two objects is physically larger or smaller, a task that requires the model to disentangle projected size on the 2D image from true 3D volume. \ding{173} \textbf{Shape Analysis:} We define a \textit{flatness score} as the ratio of the smallest dimension to the largest dimension of the object's 3D bounding box. A score close to zero indicates a highly anisotropic shape (e.g., flat like a plate or elongated like a pole). This metric enables us to ask which of two objects is "flatter" or "more elongated," testing a nuanced understanding of 3D form.

\begin{table}[t!]
\centering
\caption{Distribution of questions in the PanoEnv-QA dataset across major categories and question types.}
\vspace{-4pt}
\label{tab:dataset_stats}
\setlength{\tabcolsep}{16pt} 
\resizebox{\linewidth}{!}{%
\begin{tabular}{lrr}
\toprule
\textbf{Major Category} & \textbf{\# Questions} & \textbf{Percentage} \\
\midrule
Attribute Comparison & 2,975 & 20.1\% \\
Distance Estimation & 2,975 & 20.1\% \\
Relative Spatial Positioning & 2,975 & 20.1\% \\
Environment Identification & 2,965 & 20.0\% \\
View Source Identification & 2,937 & 19.8\% \\
\midrule
\textbf{Total} & \textbf{14,827} & \textbf{100\%} \\
\bottomrule
\end{tabular}
}
\vspace{-12pt}
\end{table}

\begin{figure*}[h!]
    \centering
    \includegraphics[width=\textwidth,height=0.36\textheight]{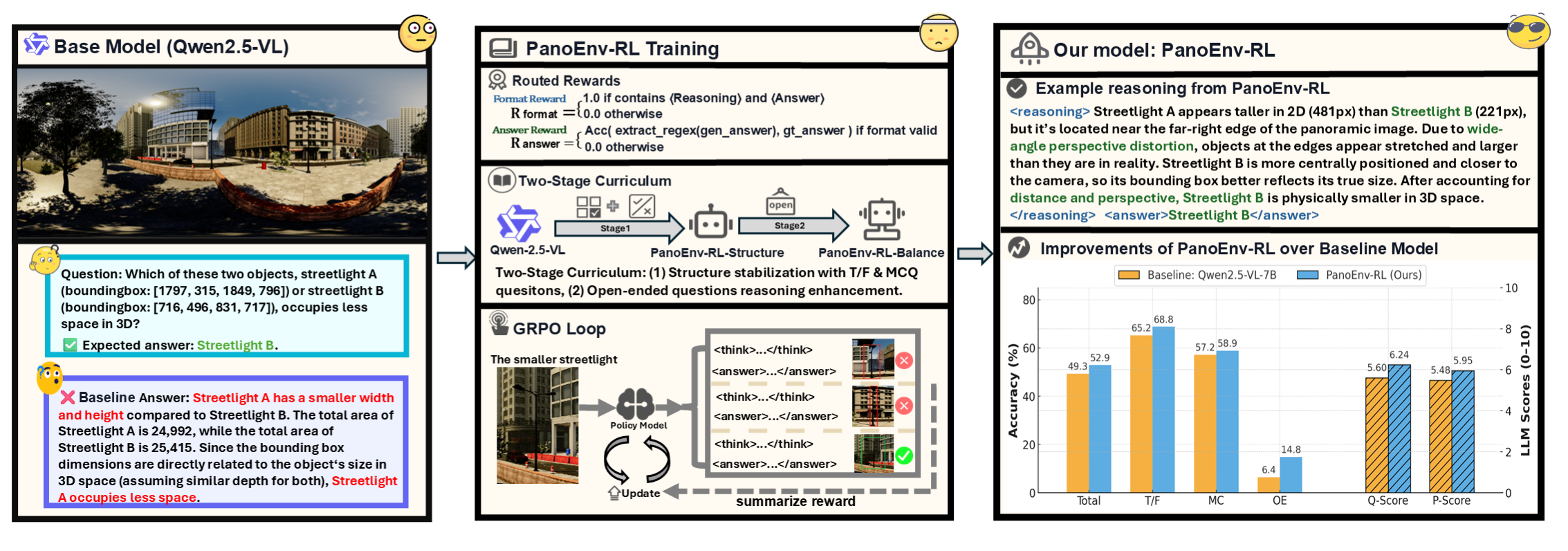}
    \vspace{-20pt}
    \caption{Overview of our framework, including GRPO sampling, routed reward computation, and two-stage curriculum updates.}
    \label{fig:rl_pipeline}
    % \vspace{-12pt}
\end{figure*}

% ---------------- 3.2 RL Framework ----------------
\subsection{3D-Aware RL Post-Training}
\label{sec:rl_method}

To address the significant performance gaps identified in our baseline benchmarks (Sec.~\ref{sec:benchmarking}), we employ a 3D-aware Reinforcement Learning (RL) post-training framework, as illustrated in Fig.~\ref{fig:rl_pipeline}. This approach allows the model to explore the output space and learn from a reward signal derived directly from our dataset's 3D ground truth. We utilize Group Relative Policy Optimization (GRPO)~\cite{shao2024deepseekmath}, a PPO~\cite{schulman2017proximal} variant optimized for language generation, to fine-tune the Qwen2.5-VL-Instruct model.

\subsubsection{Group Relative Policy Optimization (GRPO)}

GRPO is an advancement over standard PPO for language tasks . Instead of comparing a generated policy to a fixed reference policy, GRPO estimates the advantage by comparing responses within a group sampled from the current policy. For each input prompt $s$, we generate $K$ candidate responses $\{a_1, ..., a_K\}$ from our policy $\pi_{\theta}$. The advantage $A(s, a_k)$ is computed relative to the group's mean reward, which serves as the baseline:
\begin{equation}
\setlength{\abovedisplayskip}{3pt}
\setlength{\belowdisplayskip}{3pt}
A(s, a_k) = R(s, a_k) - \frac{1}{K}\sum_{i=1}^{K} R(s, a_i)
\end{equation}
This group-relative advantage estimation provides a stable training signal. The policy is then updated using the PPO clipped surrogate objective $\mathcal{L}^{\text{GRPO}}$~\cite{schulman2017proximal}, while a Kullback-Leibler (KL) divergence penalty $D_{\text{KL}}(\pi_{\theta} || \pi_{\text{ref}})$ against the original pre-trained policy $\pi_{\text{ref}}$ is added to prevent catastrophic forgetting. The final objective is:
\begin{equation}
\setlength{\abovedisplayskip}{3pt}
\setlength{\belowdisplayskip}{3pt}
\mathcal{L}_{\text{total}} = \mathcal{L}^{\text{GRPO}} - \beta \cdot D_{\text{KL}}(\pi_{\theta} || \pi_{\text{ref}})
\end{equation}
where $\beta$ is a small coefficient controlling the KL penalty.

\subsubsection{Multi-faceted Reward System}

The core of our method is a sophisticated, multi-faceted reward system that leverages the ground-truth annotations from our PanoEnv-QA dataset. The total reward $R(s, a)$ is a weighted combination of accuracy and format correctness~\cite{wei2022chain}, heavily prioritizing correctness:
\setlength{\abovedisplayskip}{3pt}
\setlength{\belowdisplayskip}{3pt}
\begin{equation}
R(s, a) = w_{\text{acc}} \, R_{\text{acc}}(a, a^*) + w_{\text{fmt}} \, R_{\text{fmt}}(a)
\end{equation}
where $a^*$ is the ground-truth answer, $R_{\text{acc}}$ and $R_{\text{fmt}}$ are the accuracy and format reward components, respectively. We set $w_{\text{acc}} = 0.9$ and $w_{\text{fmt}} = 0.1$ to heavily prioritize answer correctness.

\noindent \textbf{Format Reward ($R_{\text{fmt}}$).}
A binary reward enforcing output structure: $R_{\text{fmt}}=1.0$ if the response strictly follows \texttt{<Reasoning>...</Reasoning> <Answer>...</Answer>} (verified via regex), and 0.0 otherwise. This promotes a clear separation between reasoning and final answer.

\begin{table*}[h!]
\centering
\caption{Overall performance and per-type accuracy breakdown of 14 baseline VLMs. LLM scores (Q-Score, P-Score) are on a 0-10 scale.}
\vspace{-4pt}
\label{tab:main_benchmark_results}
\setlength{\tabcolsep}{12pt}
\renewcommand{\arraystretch}{0.85}
\resizebox{\textwidth}{!}{%
\begin{tabular}{l|ccc|ccc}
\toprule
& \multicolumn{3}{c|}{\textbf{Overall Performance}} & \multicolumn{3}{c}{\textbf{Accuracy Breakdown by Type (\%)}} \\
\midrule
\textbf{Model} & \textbf{Acc. (\%)} $\uparrow$ & \textbf{Q-Score (0-10)} $\uparrow$ & \textbf{P-Score (0-10)} $\uparrow$ & \textbf{True/False} & \textbf{Multiple Choice} & \textbf{Open-Ended} \\
\midrule
InternVL2.5-8B~\cite{chen2024expanding} & 43.64 & 5.32 & 5.60 & 63.59 & 48.29 & 3.00 \\
InternVL2.5-4B & 40.39 & 5.16 & 5.51 & \cellcolor{orange!30}\textbf{65.53} & 40.83 & 2.95 \\
InternVL2.5-26B & 47.07 & \cellcolor{orange!30}\textbf{5.61} & \cellcolor{orange!15}5.61 & 64.51 & 54.33 & 3.44 \\ \midrule
DeepSeek-VL2-Tiny~\cite{wu2024deepseek} & 31.69 & 4.55 & 4.96 & 55.67 & 30.45 & 0.16 \\
DeepSeek-VL2-Small & 35.27 & 5.02 & \cellcolor{orange!30}\textbf{5.64} & 55.78 & 35.29 & 5.57 \\
DeepSeek-VL2-Base & 38.86 & 5.24 & 5.54 & 57.30 & 40.36 & \cellcolor{orange!30}\textbf{8.36} \\  \midrule
Llama3.2-11B-Vision~\cite{dubey2024llama} & 30.71 & 4.49 & 4.94 & 57.03 & 26.91 & 2.30 \\ \midrule
LLaVA-1.5-7B~\cite{liu2024improved} & 30.12 & 4.25 & 4.59 & 50.92 & 29.80 & 0.83 \\
LLaVA-v1.6-Mistral-7B~\cite{li2024llava} & 23.62 & 3.15 & 4.78 & 45.24 & 19.06 & 3.93 \\
LLaVA-v1.6-Vicuna-13B & 18.22 & 4.00 & 5.15 & 49.60 & 7.01 & 1.32 \\ \midrule
Qwen2.5-VL-3B~\cite{bai2025qwen2} & 34.54 & 4.41 & 4.31 & 53.29 & 36.11 & 3.44 \\
Qwen2.5-VL-7B & \cellcolor{orange!30}\textbf{49.34} & \cellcolor{orange!15}5.60 & 5.48 & \cellcolor{orange!15}65.19 & \cellcolor{orange!30}\textbf{57.24} & 6.39 \\
Qwen2.5-VL-32B & 42.70 & 5.02 & 4.92 & 62.47 & 44.96 & \cellcolor{orange!30}8.36 \\
Qwen3-VL-8B~\cite{yang2025qwen3} & \cellcolor{orange!15}\textbf{47.91} & \cellcolor{orange!15}5.60 & 5.35 & 62.85 & \cellcolor{orange!15}55.24 & \cellcolor{orange!15}7.70 \\
\midrule
Average & 36.72 & 4.82 & 5.17 & 55.98 & 37.56 & 4.26 \\
\bottomrule
\end{tabular}}
\vspace{-12pt}
\end{table*}
\noindent \textbf{Accuracy Reward ($R_{\text{acc}}$).}
This component measures the correctness of the content within the \texttt{<Answer>...</Answer>} tag. We employ an automatic routing mechanism that selects one of five specialized reward strategies based on the question's pre-annotated type as follows: 
\textbf{\ding{172} Strategy A (\texttt{yes\_no}):} For \texttt{true\_false} questions. Performs a strict, case-insensitive string match (e.g., "yes" matches "Yes", but not "Yes, it is"). Assigns a binary reward (1.0 or 0.0).
\textbf{\ding{173} Strategy B (\texttt{mcq}):} For \texttt{multiple\_choice} questions. A smart subject extraction logic is used: if the answer is a long sentence, it extracts the subject (text before the first verb/preposition); otherwise, it uses the short answer directly. This extracted entity is then normalized (removing articles, punctuation, and case) and compared to the normalized ground truth for a 1.0 or 0.0 reward.
\textbf{\ding{174} Strategy C (\texttt{distance}):} For \texttt{distance} questions. A numerical parser extracts values and their units. All units (e.g., "cm", "km", "ft") are automatically converted to meters. The reward is scaled based on the relative error: 1.0 for $\le$10\% error, 0.5 for $\le$20\% error, and 0.0 otherwise.
\textbf{\ding{175} Strategy D (\texttt{spatial}):} For \texttt{spatial} relationship questions. This strategy parses the answer for directional keywords (and their synonyms, e.g., "above" = "over") across three independent axes (front/back, left/right, up/down). Compares the sets of keywords for each dimension independently. The reward is the fraction of correctly identified dimensions (e.g., matching 2 of 3 required axes, like "behind and right" when the answer is "behind, right, and above", yields a reward of 0.67).
\textbf{\ding{176} Strategy E (\texttt{counting}):} For \texttt{counting} questions. An exact numerical match is required. The parser supports both digits ("3") and textual numbers ("three"), converting them to a standard value for comparison (1.0 or 0.0 reward). This fine-grained, routed reward system provides a much more accurate and targeted learning signal than a single, generic reward function, directly leveraging the physical ground truth inherent in our dataset.

\subsubsection{Two-Stage Curriculum Strategy}
\label{sec:curriculum}

Training RL on mixed-format panoramic QA is unstable due to heterogeneous reward signals and the high entropy of open-ended generation. We therefore adopt a \emph{two-stage curriculum}, consistent with curriculum learning principles~\cite{bengio2009curriculum} and recent practices in RL fine-tuning for LLMs~\cite{parashar2025curriculum,narvekar2020curriculum}. Our design is motivated by findings in curriculum RL, where structuring training as a progression from simpler to more challenging subtasks reduces exploration complexity and improves learning stability~\cite{florensa2017reverse, matiisen2019teacher,xi2024training}.
\textbf{\ding{172} Stage 1: Structured Pretraining.}
We first train solely on structured questions (MCQ and T/F). These low-entropy tasks yield reliable rewards and allow the policy to quickly acquire formatting discipline and stable short-form reasoning. \textbf{\ding{173} Stage 2: Mixed Open-Ended Training.}
We then fine-tune the model on a balanced mixture of structured and open-ended questions. With formatting and basic reasoning already established, the model can focus on improving open-ended spatial reasoning without suffering catastrophic forgetting.
Overall, this curriculum stabilizes optimization and enables efficient transfer of structured skills to challenging panoramic 3D reasoning, as validated by our ablation.

\begin{table*}[t!]
\centering
\caption{
Comprehensive comparison of our GRPO-trained models (\textbf{Ours}) against 14 state-of-the-art baselines on the PanoEnv-QA test set. 
All our models are fine-tuned on Qwen2.5-VL-7B using LoRA.}
% \textbf{GRPO-Balanced} achieves the best performance across Total Accuracy, Open-Ended Accuracy, and both LLM-based quality scores, validating the effectiveness of our two-stage curriculum learning framework.
\vspace{-4pt}
\label{tab:rl_main_results}
\setlength{\tabcolsep}{4pt} 
\resizebox{\textwidth}{!}{%
\begin{tabular}{l|c|ccc|c|c|cc}
\toprule
\textbf{Model} & \textbf{Total Acc. (\%)} $\uparrow$ & \textbf{T/F (\%)} & \textbf{MC (\%)} & \textbf{OE (\%)} & \textbf{T/F+MCQ Acc. (\%)} & \textbf{Params} & \textbf{Q-Score (0--10)} $\uparrow$ & \textbf{P-Score (0--10)} $\uparrow$ \\
\midrule
Qwen2.5-VL-7B (Base) & 49.34 & 65.19 & 57.24 & 6.39 & 60.46 & 7B & 5.60 & 5.48 \\
Qwen2.5-VL-32B & 42.70 & 62.47 & 44.96 & 8.36 & 52.46 & 32B & 5.02 & 4.92 \\
\midrule
\textbf{GRPO-Balanced (Ours)} & \textbf{52.93} & \textbf{68.78} & \textbf{58.90} & \textbf{14.83} & \textbf{62.89} & 7B & \textbf{6.24} & \textbf{5.95} \\
\bottomrule
\end{tabular}}
\vspace{-12pt}
\end{table*}

\section{Experiments and Analysis}
\label{sec:experiments}

Our experiments evaluate (1) the quality of the PanoEnv-QA dataset, (2) the zero-shot capability of existing VLMs, and (3) the effectiveness of our 3D-aware RL post-training framework. The section concludes with detailed ablation studies validating every design choice.

% -------------------------
% 1. DATASET ANALYSIS
% -------------------------
\subsection{Dataset Analysis}
\label{sec:dataset_analysis}

Following the methodology described in Section~\ref{sec:dataset_construction}, we generated the PanoEnv-QA dataset. The final dataset is a large-scale collection comprising \textbf{14,827} high-quality QA pairs derived from \textbf{595} panoramic scenes spanning \textbf{60} diverse virtual environments. On average, each scene contributes around 25 QA pairs.

\noindent \textbf{\ding{172} Question distribution and balance:}
A key design goal of PanoEnv-QA is to provide a balanced testbed for spatial reasoning. As shown in Table~\ref{tab:dataset_stats}, each of the five core categories constitutes roughly 20\% of the dataset, ensuring that no single skill dominates overall model performance.

\noindent \textbf{\ding{173} Answer characteristics and diversity:}
The dataset contains \textbf{1,894} unique answers with an average length of 10.9 characters. Answer formats range from binary (e.g., ``No''), quantitative (e.g., ``About 4.2 meters''), and object names (e.g., ``The carousel'') to full-sentence spatial descriptions.  
Importantly, the Yes/No ratio across all binary questions (45.3\% vs.~54.7\%) prevents majority-class shortcuts.
Overall, PanoEnv-QA is balanced, diverse, and challenging, making it a robust benchmark for holistic spatial reasoning.

% -------------------------
% 2. BASELINE BENCHMARKING
% -------------------------
\subsection{Benchmarking Baseline Models}
\label{sec:benchmarking}

We evaluate 14 SOTA VLMs—including Qwen-VL, InternVL, and DeepSeek-VL—on a 3,040-sample test set.

\noindent \textbf{Evaluation Metrics.}
We report:
1) \textbf{Accuracy}: a strict rule-based score using specialized parsers;  
2) \textbf{Qwen-Score (Q-Score)}~\cite{yang2025qwen3};  
3) \textbf{Prometheus-Score (P-Score)}~\cite{kim2024prometheus}.  

\noindent \textbf{Results and analysis.}
As seen in Table~\ref{tab:main_benchmark_results}, zero-shot performance is very low. The strongest baseline, Qwen2.5-VL-7B, reaches only \textbf{49.34\%} accuracy, and open-ended accuracy collapses to \textbf{6.39\%}.  
These failures strongly motivate our 3D-aware RL post-training.

\subsection{RL Post-Training Results}
\label{sec:rl_results}

We fine-tune Qwen2.5-VL-7B-Instruct using GRPO to improve the weak Open-Ended (OE) reasoning observed in zero-shot VLMs. All models are trained for two epochs with group size $K=4$, LoRA~\cite{hu2022lora} applied to the language decoder, and a frozen vision encoder.

\noindent \textbf{Two-Stage Curriculum.}
Our RL procedure follows the two-stage curriculum introduced in Sec.~\ref{sec:curriculum}:
\ding{172} \textbf{GRPO-Structured (Stage~1):} trained solely on T/F and MCQ questions using aggressive hyper-parameters to rapidly learn format adherence and discrete decision-making.  
\ding{173} \textbf{GRPO-Balanced (Stage~2, Ours):} initialized from Stage~1 and trained on a balanced mix of all OE questions with an equal number of structured ones, using conservative hyper-parameters to stabilize optimization while improving OE reasoning.
\begin{figure}[t!]
    \centering
    \includegraphics[width=0.98\linewidth]{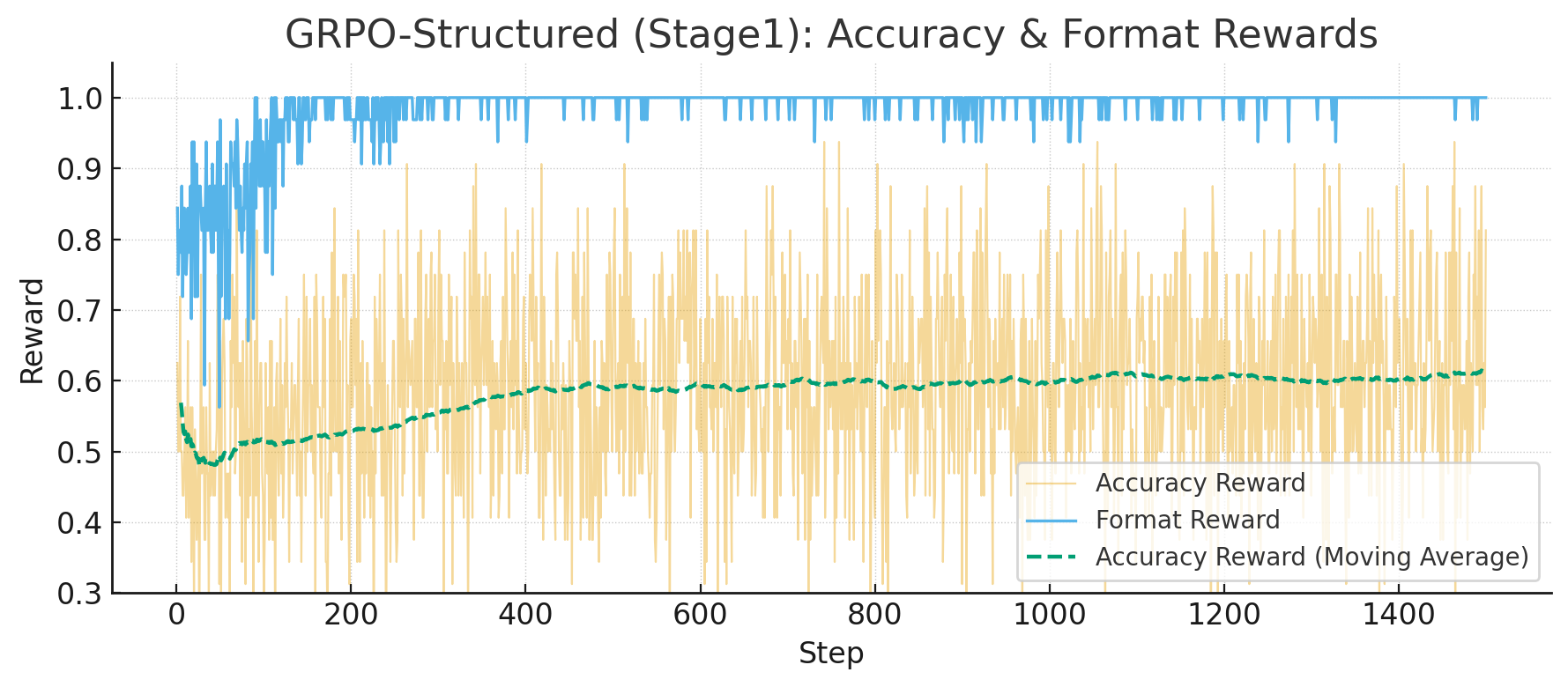}
    % \vspace{1.2em}
    \includegraphics[width=0.98\linewidth]{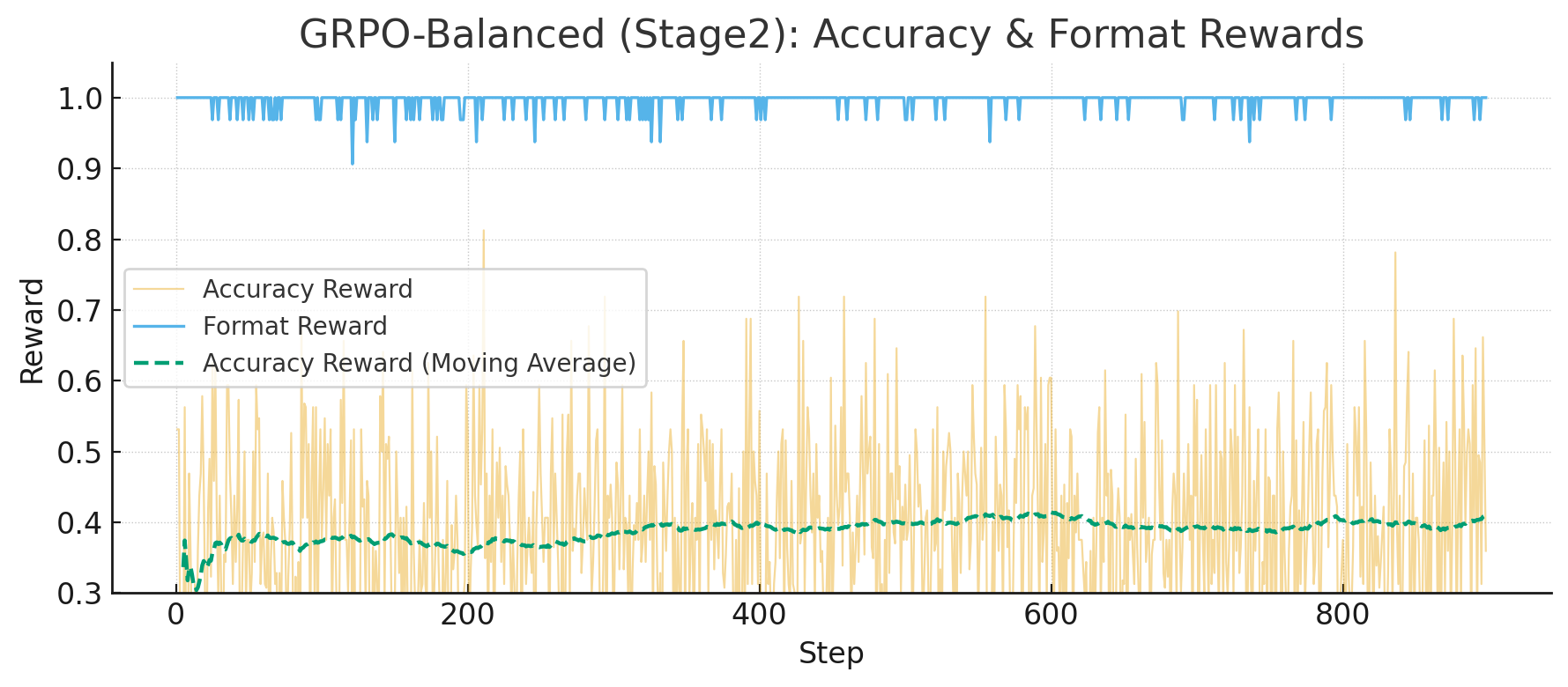}
    \vspace{-4pt}
    \caption{\textbf{Training dynamics of our two-stage GRPO curriculum.}
    Stage~1 quickly learns output format and structured decision-making; Stage~2 inherits this and focuses on improving OE reasoning under balanced training.}
    \label{fig:rl_curves}
\end{figure}
As illustrated in Fig.~\ref{fig:rl_curves}, Stage~1 quickly pushes the format reward close to $1.0$ and steadily increases the accuracy reward (moving average rising from approximately 0.5 to above 0.6), showing that the model rapidly masters structured decisions under strong supervision. In Stage~2, the format reward remains saturated from the beginning, while the accuracy reward continues to improve from a lower starting point, indicating that the model now focuses on harder OE reasoning without sacrificing the learned output structure.
\noindent \textbf{Main Results.}
As shown in Table~\ref{tab:main_benchmark_results} and Table~\ref{tab:rl_main_results}, our \textbf{GRPO-Balanced} model, representing our full \textbf{PanoEnv-RL} framework, achieves the best overall performance on PanoEnv-QA with \textbf{52.9\%} accuracy (vs.\ 49.3\% zero-shot). It also attains the highest LLM-judge scores (Q-Score 6.24, P-Score 5.95) and boosts OE accuracy from \textbf{6.4\%} to \textbf{14.8\%}. Combined with the reward trends in Fig.~\ref{fig:rl_curves}, these results demonstrate that the two-stage GRPO curriculum effectively enhances geometry-grounded reasoning in 360° scenes while maintaining structured outputs.

\subsection{Ablation Studies}
\label{sec:ablation}
We analyze five controlled variants to assess the effects of curriculum design and OE integration (Table~\ref{tab:ablation_unified}). The two-stage setup is crucial: \textbf{GRPO-Balanced} outperforms \textbf{GRPO-OneStage} (50.8\%) and \textbf{Reverse} (50.9\%), achieving the best overall (52.9\%) and OE (14.8\%) results. Structured→mixed training yields a more stable optimization path than training on all tasks simultaneously. \textbf{GRPO-Structured} performs well on T/F and MCQ but collapses on OE (5.7\%). Adding balanced OE data in Stage~2 restores these abilities without harming structured accuracy. In contrast, \textbf{GRPO-OE} moderately improves OE (13.2\%) but weakens structured skills. Overall, \textbf{GRPO-Balanced} achieves the best stability and performance across all tasks.

\begin{table}[t]
\centering
\caption{Unified ablation results across all variants.}
\label{tab:ablation_unified}
\vspace{0.2em}
\resizebox{\linewidth}{!}{
\begin{tabular}{l|c|ccc|cc}
\toprule
\textbf{Model Variant} & \textbf{Total} & \textbf{T/F} & \textbf{MCQ} & \textbf{OE} & \textbf{Q-Scr} & \textbf{P-Scr} \\
 & (\%) & (\%) & (\%) & (\%) & (0--10) & (0--10) \\
\midrule
Baseline (Qwen2.5-VL-7B) & 49.3 & 65.2 & 57.2 & 6.4 & 5.60 & 5.48 \\
GRPO-OneStage (All-in-One) & 50.8 & 67.6 & 56.7 & 11.8 & 6.22 & 5.89 \\
GRPO-Structured (Structured-only) & 52.3 & \textbf{69.5} & \textbf{60.9} & 5.7 & 6.14 & 5.94 \\
GRPO-OE (OE-only) & 48.6 & 66.6 & 52.3 & 13.2 & 5.56 & 5.47 \\
GRPO-Reverse (OE$\rightarrow$Mixed) & 50.9 & 69.3 & 57.8 & 7.0 & 6.16 & 5.91 \\
\midrule
\textbf{GRPO-Balanced (Struct$\rightarrow$Mixed)} & \textbf{52.9} & 68.8 & 58.9 & \textbf{14.8} & \textbf{6.24} & \textbf{5.95} \\
\bottomrule
\end{tabular}}
\vspace{-0.3em}
\end{table}

\section{Conclusion}
\label{sec:conclusion}

In this paper, we addressed the challenge of 3D spatial reasoning in Vision–Language Models (VLMs) on 360° Equirectangular Projection (ERP) images, caused by geometric distortion and limited 3D supervision. We made three main contributions. First, we introduced \textbf{PanoEnv-QA}, a large-scale VQA benchmark with 14.8K questions across five categories, built from synthetic 3D environments and grounded in precise, verifiable 3D annotations for reliable evaluation and training. Second, benchmarking 14 state-of-the-art VLMs revealed major weaknesses in 3D understanding—achieving only 49.34\% overall and 8.36\% on open-ended (OE) questions. Third, we proposed a 3D-aware reinforcement learning framework based on Group Relative Policy Optimization (GRPO), featuring a ground-truth-guided reward with five geometry-aware strategies and a two-stage curriculum to ensure stable learning and prevent forgetting. Our 7B model achieves state-of-the-art results on PanoEnv-QA (52.93\% total accuracy), with OE accuracy rising from 6.39\% to \textbf{14.83\%} (+132\% relative), outperforming 32B models. These results demonstrate that PanoEnv and our framework effectively enhance 3D spatial intelligence in VLMs for omnidirectional perception.

\noindent \textbf{Limitations and Future Work.}
While PanoEnv is built on high-fidelity synthetic data, the synthetic-to-real gap remains a challenge. Future work could focus on adapting our RL framework to real-world 360° datasets, which often have noisy or incomplete GT. Furthermore, extending this spatial reasoning framework to temporal tasks in panoramic video remains a complex and promising direction.
\clearpage

{
    \small
    \bibliographystyle{ieeenat_fullname}
    \bibliography{main}
}

\clearpage
\section{Supplementary Material}

\subsection{More Related Works}
\label{ssec:more_related_works}

\noindent \textbf{RL for Multimodal and Spatial Reasoning.}
Supervised fine-tuning (SFT) is the standard training method for MLLMs, but its reasoning processes can be rigid~\cite{dong2024abilities}. RL offers a promising alternative, as demonstrated by pioneering works~\cite{guo2025deepseek,ouyang2022training}, such as Vision-R1~\cite{huang2025vision}, which converts visual inputs into structured reasoning chains. 

Recently, this R1-style reinforcement learning paradigm has been rapidly extended beyond 2D vision to 3D spatial understanding and embodied agent domains. Recent breakthroughs, including 3D-R1~\cite{huang20253d}, Robot-R1~\cite{kim2025robot}, VLN-R1~\cite{qi2025vln}, and Nav-R1~\cite{liu2025nav}, have collectively demonstrated that RL fine-tuning significantly enhances complex geometric reasoning, physical interaction, and spatial navigation in embodied scenes. 

In the panoramic VQA domain specifically, 360-R1~\cite{zhang2025towards} also applies an RL framework with a reward function based on semantic similarity. However, previous methods rely heavily on other LLMs or human annotations to obtain answers or CoT, which inevitably introduces biases or hallucinations. The core innovation in our PanoEnv research is the use of geometric ground truth from TartanAir~\cite{wang2020tartanair} to \textbf{programmatically generate correct textual answers}, which then serve as the basis for reward calculation. During RL training, the reward is based on the correspondence between the model's output and this physically-grounded ground truth text. This ensures the learning signal originates entirely from the objective physical world, providing a more direct and rigorous incentive for the model to learn true 3D spatial structures.

\subsection{Dataset Construction Details}

\subsubsection{Object Filtering and Sampling Heuristics}
\label{sssec:filtering}

\vspace{4pt}
\noindent \textbf{Objective.} Before question generation, we implement a rigorous object filtering and sampling pipeline to ensure the high quality and physical validity of the generated QA pairs, mitigating the geometric distortions inherent in Equirectangular Projection (ERP) images.

\vspace{4pt}
\noindent \textbf{Ground Truth Formulation.} The filtering mechanism operates on three hierarchical levels:

\vspace{4pt}
\noindent \textbf{1. Size and Aspect Ratio Constraints.} 
Objects are retained only if their 2D bounding box area in the ERP image satisfies $\text{Area} \ge 900$ pixels, with both $\text{width} \ge 25$ and $\text{height} \ge 25$ pixels. Highly distorted or degenerate masks are removed by thresholding the aspect ratio to strictly $< 5$.

\vspace{4pt}
\noindent \textbf{2. Semantic Exclusion.} 
We systematically exclude background surfaces with low QA discrimination (e.g., \textit{sky}, \textit{ground}, \textit{wall}), extremely thin objects prone to detection errors (e.g., \textit{wire}), and unstable semantic debris (e.g., \textit{rubble}).

\vspace{4pt}
\noindent \textbf{3. Containment Filtering.} 
For relational and comparative questions, we exclude object pairs with substantial bounding box overlap to prevent visual ambiguity. Two objects are deemed unsuitable for comparison if their bounding box overlap ratio exceeds $0.90$.

\subsubsection{Robust Depth Profiling}
\label{sssec:robust_depth}

As mentioned in the main text, relying solely on a single median depth value can lead to significant errors for ``thick'' objects or objects exhibiting severe partial occlusion. Therefore, we utilize the extracted point cloud to compute a robust depth profile based on percentiles ($p_{n}$) and the Inter-Quartile Range ($IQR = p_{75} - p_{25}$).

\vspace{4pt}
\noindent \textbf{Effective Depth for Comparison.} 
An object is defined as having a significant physical thickness along the camera ray if its depth variation satisfies:
\begin{equation}
    IQR > \max(0.6, 0.15 \cdot p_{50})
\end{equation}
For objects meeting this condition, using the median depth ($p_{50}$) may misrepresent their closest visible surface. Instead, we dynamically route the depth metric to the near-end quantile ($p_{20}$) to serve as the effective depth for distance-based QA comparisons.
    
\vspace{4pt}
\noindent \textbf{Distance Similarity Thresholds.} 
When generating ``True/False'' questions regarding whether two objects are at a similar distance, we evaluate both absolute median differences and interval overlaps. Two objects, $A$ and $B$, are considered to be at a similar depth if the Jaccard similarity of their inter-quartile depth intervals exceeds $0.30$:
\begin{equation}
    \frac{|[p_{25}^{A}, p_{75}^{A}] \cap [p_{25}^{B}, p_{75}^{B}]|}{|[p_{25}^{A}, p_{75}^{A}] \cup [p_{25}^{B}, p_{75}^{B}]|} > 0.30
\end{equation}
Alternatively, they are considered similar if the absolute difference between their median depths is exceptionally small:
\begin{equation}
    |p_{50}^{A} - p_{50}^{B}| < \max(0.5, 0.10 \cdot \min(p_{50}^{A}, p_{50}^{B}))
\end{equation}

\subsubsection{Data Quality Assurance via Human Verification}
\label{sssec:human_verification}
To rigorously validate the semantic soundness of our programmatically generated ground truth, we conducted a human evaluation study on a randomly sampled subset of the PanoEnv-QA dataset. Independent human annotators verified the correctness of the generated answers against the visual evidence in the ERP images. The human verification yielded a \textbf{96\% accuracy rate}, confirming that our physics-derived geometric ground truth is highly reliable, semantically sound, and strongly aligns with human perception of 3D spaces.

\subsection{Prompt Designs for Generation and Evaluation}
\label{ssec:prompt_designs}

The design of the prompt is critical for both the GRPO exploration phase and the final model evaluation. In this section, we detail the unified templates used for eliciting model responses, as well as the strict scoring prompts utilized by the LLM-as-a-judge.

\subsubsection{Generation and Training Prompts}
\label{sssec:generation_prompts}

To encourage the model to output transparent Chain-of-Thought (CoT) reasoning before providing the final answer, we strictly enforce a unified prompt structure across all baseline evaluations and our RL training. 

The complete prompt is constructed dynamically by concatenating a Base Prompt with a Task-Specific Suffix depending on the question type.

\vspace{4pt}
\noindent \textbf{Base Prompt (Shared across all tasks):}
\begin{quote}
\ttfamily\small
You are an expert in analyzing 360$^\circ$ \\
panoramic (ERP) images (2560x1280). \\
Analyze the image carefully and focus on \\
the specific objects mentioned in \\
bounding boxes. \\
\\
\{Question\} \\
\\
Provide your reasoning based on the \\
panoramic scene within <Reasoning> tags, \\
then give your final answer within \\
<Answer> tags.
\end{quote}

\vspace{4pt}
\noindent \textbf{Task-Specific Format Constraints:} \\
To ensure robust parsing for our routed reward system, one of the following rule suffixes is appended to the base prompt based on the question category:

\vspace{4pt}
\noindent \textbf{1. True/False Questions:}
\begin{quote}
\ttfamily\small
This is a YES/NO question. Your <Answer> \\
must be EXACTLY "Yes" or "No" \\
(case-sensitive, no extra words). \\
Example: \\
<Reasoning>...</Reasoning>\\
<Answer>Yes</Answer>
\end{quote}

\vspace{4pt}
\noindent \textbf{2. Multiple-Choice Questions:}
\begin{quote}
\ttfamily\small
This is a MULTIPLE CHOICE question. \\
Your <Answer> must be EXACTLY one of \\
the provided options. \\
Example: \\
<Reasoning>...</Reasoning>\\
<Answer>AbandonedCable</Answer>
\end{quote}

\vspace{4pt}
\noindent \textbf{3. Open-Ended Questions:}
\begin{quote}
\ttfamily\small
This is an OPEN-ENDED question. \\
Your <Answer> should be concise and \\
direct (under 20 words). \\
Example: \\
<Reasoning>...</Reasoning>\\
<Answer>The building is behind and to \\
the right of and below the truck</Answer>
\end{quote}

\subsubsection{LLM-as-a-Judge Evaluation Prompts}
\label{sssec:eval_prompts}

In addition to the strict rule-based accuracy metrics, we employ an LLM-as-a-judge approach (e.g., QwenScore and PrometheusScore) to evaluate the semantic correctness and reasoning quality of the models' outputs. To ensure consistency and transparency, we use a highly structured prompt that explicitly defines the scoring rules (0-10) for each specific question type.

\vspace{4pt}
\noindent \textbf{1. System Prompt (Evaluator Rules):}
\begin{quote}
\ttfamily\small
You are an evaluator. Compare answers and \\
give a score 0-10. \\
\\
Rules:\\
- YES/NO: yes/true/1 = YES, no/false/0 = NO. \\
  Same meaning -> 10, different -> 0. \\
- Multiple choice: output a single option only. \\
  Ignore articles/case/punctuation. \\
  Match -> 10, else -> 0. \\
- Numeric (distance, e.g. "About X meters"): \\
  extract numbers. If no valid number -> 0. \\
  * <=10\% relative error -> 10 \\
  * <=20\% -> 5-9 \\
  * >20\% -> 0 \\
- Numeric (counting, e.g. "3 different views"): \\
  use only the integer. \\
  * Exact match -> 10 \\
  * Otherwise -> 0 \\
- Spatial: compare direction words (left/right, \\
  front/behind, above/below). \\
  If any axis is opposite (e.g. left vs right), \\
  score -> 0. Otherwise, let N = axes used in \\
  reference, C = correctly matched axes; \\
  score $\approx$ 10 * C / N. \\
- Open-ended: judge semantic match. \\
  * All key info correct -> 10 \\
  * Most info correct -> 8-9 \\
  * Partially correct -> 6-7 \\
  * Mostly wrong -> 0-5 \\
\\
IMPORTANT: End response with "Score: X" \\
where X is 0-10.
\end{quote}

\vspace{4pt}
\noindent \textbf{2. User Prompt (Evaluation Instance):} \\
During evaluation, the test samples are dynamically formatted into the following template and passed to the LLM judge:

\begin{quote}
\ttfamily\small
Q: \{question\} \\
Type: \{question\_type\} \\
Reference: \{expected\_answer\} \\
Answer: \{model\_answer\} \\
\\
Give score 0-10. Must end with "Score: X":
\end{quote}

\subsection{Training Details and Hyperparameters}
\label{ssec:training_details}

\noindent \textbf{Hardware \& Optimization.} Models are trained on 4 $\times$ NVIDIA B200 GPUs with \texttt{bfloat16}, DeepSpeed ZeRO-2, and Flash Attention 2. For reproducibility, training is also feasible on 80GB A100 GPUs by halving the group size ($K=2$) and per-device batch size. Hyperparameters are detailed in Table~\ref{tab:hyperparameters}.

\begin{table}[h]
\centering
\caption{Hyperparameter configurations for GRPO fine-tuning.}
\vspace{-4pt}
\label{tab:hyperparameters}
\resizebox{\linewidth}{!}{%
\renewcommand{\arraystretch}{1.1}
\begin{tabular}{lc}
\toprule
\textbf{Hyperparameter} & \textbf{Value} \\
\midrule
Base Model & Qwen2.5-VL-7B-Instruct \\
Tuning Method & LoRA (Language Decoder only) \\
LoRA Rank ($r$) / Alpha ($\alpha$) & 16 / 32 \\
LoRA Dropout & 0.05 \\
Group Size ($K$) & 4 \\
Training Epochs & 2 \\
Peak Learning Rate & 5e-6 \\
Warmup Ratio & 0.05 \\
Global Batch Size & 32 \\
KL Coefficient ($\beta$) & 0.01 \\
Max Completion Length & 256 \\
Max Gradient Norm & 10.0 \\
Reward Weights (Acc, Fmt) & (0.90, 0.10) \\
\bottomrule
\end{tabular}%
}
\vspace{-6pt}
\end{table}

\subsection{Comparison with Existing Benchmarks}
\label{ssec:benchmark_comparison}

To further clarify the unique positioning of PanoEnv-QA, we provide a detailed comparison with recent panoramic reasoning benchmarks, such as OSR-Bench and OmniVQA, in Table~\ref{tab:benchmark_comparison}. Specifically, PanoEnv uniquely focuses on recovering 3D metric and physical relationships from 2D pixels using precise simulation engine ground truth.

\begin{table}[h]
\centering
\caption{Comprehensive comparison between PanoEnv and existing panoramic benchmarks.}
\vspace{-4pt}
\label{tab:benchmark_comparison}
\resizebox{\linewidth}{!}{%
\renewcommand{\arraystretch}{1.1}
\begin{tabular}{l|ccc}
\toprule
\textbf{Feature} & \textbf{PanoEnv (Ours)} & \textbf{OSR-Bench} & \textbf{OmniVQA} \\
\midrule
\textbf{Dataset Base} & TartanAir & Replica / DeepPano & Stan. 2D-3D-S \\
\textbf{Scenes} & 60 (Indoor \& Outdoor) & 1,526 (Indoor) & 6 (Indoor) \\
\textbf{Images} & 595 & 4,100 & 1,213 \\
\textbf{QA Pairs} & $\sim$14.8K & $\sim$153K & $\sim$4.9K \\
\midrule
\textbf{Core Focus} & \textbf{2D-to-3D Inference} & Topology & Polar Distortion \\
\textbf{GT Source} & \textbf{3D Annotations} & Cognitive Map & MLLM + Human \\
\bottomrule
\end{tabular}%
}
\vspace{-6pt}
\end{table}

Unlike OSR-Bench, which focuses on 2D topological relations and achieves its scale through extensive negative sampling, PanoEnv prioritizes reasoning depth over massive dataset scale. Our benchmark evaluates the implicit recovery of 3D physical reality (e.g., true volume and metric distance) from monocular panoramas. Furthermore, in contrast to OmniVQA which relies on LLM-generated answers, PanoEnv derives its rewards strictly from pixel-perfect geometric annotations provided by the simulation engine, thereby preventing the reinforcement of hallucinated reasoning.

\subsection{Additional Experiments: Sim-to-Real Generalization}
\label{ssec:sim_to_real}

To validate the real-world transferability of PanoEnv-RL, we conducted a zero-shot evaluation on the \textbf{OSR-Bench}, a benchmark comprised of photorealistic panoramas derived from high-fidelity scene scans.

As shown in Table~\ref{tab:osr_bench_supp}, despite being trained \textit{solely} on simulation data (TartanAir), our PanoEnv-RL (7B) consistently outperforms the Base 7B model and even surpasses the significantly larger \textbf{Qwen2.5-VL-72B} on tasks such as \textit{Object Counting} and \textit{Relative Distance}. 

\begin{table}[h]
\centering
\caption{Zero-Shot Performance on OSR-Bench (Real-World).}
\vspace{-4pt}
\label{tab:osr_bench_supp}
\resizebox{\linewidth}{!}{%
\renewcommand{\arraystretch}{1.1} 
\begin{tabular}{l|ccc}
\toprule
\textbf{Model} & \textbf{Obj. Count} & \textbf{Rel. Dist.} & \textbf{Rel. Dir.} \\
\midrule
Qwen2.5-VL-7B (Base) & 0.477 & 0.321 & 0.089 \\
Qwen2.5-VL-72B & 0.498 & 0.325 & \textbf{0.181} \\
\textbf{PanoEnv-RL (Ours)} & \textbf{0.507} & \textbf{0.371} & 0.105 \\
\bottomrule
\end{tabular}%
}
\vspace{-6pt}
\end{table}

\vspace{4pt}
\noindent \textbf{Transferability.} The results exhibit the robust transferability of our method to real-world domains. The performance gap in \textit{Relative Direction} compared to the 72B model is primarily attributed to a \textbf{2D-vs-3D mismatch}: OSR-Bench lacks verticality and evaluates flat 2D topological relations, which inherently penalizes our model's fine-grained 3D spherical predictions.

\vspace{4pt}
\noindent \textbf{Logic over Scale.} The ability to generalize directly to the real-world OSR-Bench without any domain-specific fine-tuning confirms that PanoEnv-RL acquires transferable \textbf{3D spatial reasoning} (i.e., geometric logic) rather than merely memorizing simulation data or textures. PanoEnv provides the exact physical ground truth essential to encode this reasoning core, successfully enabling robust sim-to-real transfer and proving that geometric logic can bridge the reality gap more effectively than simply scaling model parameters.

\end{document}